\documentclass[preprint,12pt,3p,authoryear]{elsarticle}
\usepackage[english]{babel}
\usepackage{graphicx}
\usepackage{epstopdf}
\usepackage{booktabs}
\usepackage{rotating}
\usepackage{listings}
\usepackage{color}
\usepackage[cmex10]{amsmath}
\usepackage{amsfonts}
\usepackage{amssymb}
\usepackage{amsthm}
\usepackage{algorithm}
\usepackage{algorithmicx}
\usepackage{algpseudocode}
\usepackage{morefloats}
\usepackage{bm}
\usepackage{setspace}
\usepackage{multirow}
\biboptions{sort}

\begin{document}

\title{Towards Clinical Diagnosis: Automated Stroke Lesion Segmentation on Multimodal MR Image Using Convolutional Neural Network}%
\author[nk]{Zhiyang~Liu}
\ead{liuzhiyang@nankai.edu.cn}
\author[hh]{Chen~Cao\corref{cor1}}
\ead{bdqjqs338@163.com}
\author[aizu,nk]{Shuxue~Ding}
\ead{sding@u-aizu.edu.jp}
\author[hh]{Tong~Han}
\ead{mrbold@163.com}
\author[nk]{Hong~Wu}
\ead{wuhong@nankai.edu.cn}
\author[tmu]{Sheng~Liu}
\ead{blueice7011@hotmail.com}

\cortext[cor1]{Corresponding author}

\address[nk]{Tianjin Key Laboratory of Optoelectronic Sensor and Sensing Network Technology, College of Electronic
Information and Optical Engineering, Nankai University, Tianjin 300071, China.}
\address[hh]{Key Laboratory for Cerebral Artery and Neural Degeneration of Tianjin, Department of Medical Imaging, Tianjin Huanhu Hospital, Nankai University, Tianjin 300350, China.}
\address[aizu]{School of Computer Science and Engineering, the University of Aizu,
Aizu-Wakamatsu, Fukushima 965-8580, Japan.}
\address[tmu]{School of Basic Medical Sciences, Tianjin Medical University, Tianjin 300071, China.}

\begin{abstract}
The patient with ischemic stroke can benefit most from the earliest possible definitive diagnosis. While the high quality medical resources are quite scarce across the globe, an automated diagnostic tool is expected in analyzing the magnetic resonance (MR) images to provide reference in clinical diagnosis. In this paper, we propose a deep learning method to automatically segment ischemic stroke lesions from multi-modal MR images. By using atrous convolution and global convolution network, our proposed residual-structured fully convolutional network (Res-FCN) is able to capture features from large receptive fields. The network architecture is validated on a large dataset of 212 clinically acquired multi-modal MR images, which is shown to achieve a mean dice coefficient of 0.645 with a mean number of false negative lesions of 1.515. The false negatives can reach a value that close to a common medical image doctor, making it exceptive for a real clinical application.
\end{abstract}


\date{}%

\maketitle

\section{Introduction}

Ischemic stroke is the most common cerebrovascular disease and one of the most frequent causes of death and disability worldwide. A patient with ischemic stroke can benefit most from the earliest possible definitive diagnosis, and imaging plays an essential role in the assessment of patients \citep{Crichton2016}. Due to its excellent soft tissue contrast, the magnetic resonance imaging (MRI) has become the modality of choice for clinical evaluation of ischemic stroke lesions. For a quantitative analysis of stroke lesion in MRI images, the expert manual segmentation is still a common approach and has been employed to compute the size, shape and volume of the stroke lesions. However, it is a tedious and time consuming task and is non-reproducible. Therefore, the development of fully automated and accurate stroke lesion segmentation method has become an active research field, but it is not easy task\citep{Ashton2003}.

Conventionally, the lesion segmentation is treated as an abnormality detection problem, where a healthy atlas is established, and the lesions are detected according to the differences in tissue appearance\citep{ Doyle2013, Gooya2011, Schmidt2012, Liu2014}. The brain appearance, however, differs from patient to patient, and the lesion may also cause deformation in brain structure. Moreover, the MRI acquired from different machines may also introduce different levels of noise and deformation of brain tissue appearance, leading to incorrect detection and segmentation. Therefore, many machine-learning methods have been proposed, where the features are learnt from massive training data, and high segmentation accuracy can be achieved. For instance,  random forest based methods were used in the literature \citep{Ellwaa2016, LeFolgoc2016,Lefkovits2016, Song2016}, which presents good performance in brain tumor segmentation by using hand-crafted features.

Note that the performance of the random-forests-based methods heavily rely on the manually annotated features. To achieve better performance, it is preferable to make the machine find the features from the data by itself. The deep learning is a machine-learning approach that uses layered hierarchical, graphical networks to extract features from data at progressively higher levels of abstraction\citep{LeCun2015}. In recent years, the deep-learning-based methods have been widely used in object classification and semantic image segmentation thanks to its recent breakthrough in convolutional neural network (CNN). The deep learning methods are originally used for image classification for daily images, such as flowers, persons, etc., and have achieved $2.3\%$ top-5 error in the classification of 1000 objects in Imagenet Large Scale Visual Recognition Challenge (ILSVRC) 2017. When applied to biomedical image classification and segmentation, the deep learning methods suffers from the lack of data. For instance, the ILSVRC 2017 dataset contains $1,431,167$ images, while the brain tumor segmentation (BraTS) challenge 2015 has only 274 patients in the training dataset and 110 patients in the testing dataset\citep{Menze2015}. In the sub-acute stroke lesion segmentation task of the ischemic stroke lesion segmentation (ISLES) 2015 challenge, the dataset is much smaller, with 28 patients in the training dataset and 36 patients in the testing dataset\citep{Maier2017}. The insufficient data limits its ability to learn features from the training data, and may also lead to over-fitting on the training data. Despite of this, many deep-learning-based methods have been proposed for brain tumor and ischemic stroke segmentation\citep{Havaei2017, Kamnitsas2016, Kamnitsas2017,Pereira2016, Randhawa2016}, and presents good performance. Note that the CNN is originally developed for image classification, one of the most popular methods is to convert the image segmentation task to a pixel-by-pixel classification, and dedicated loss functions have been designed to overcome the huge class imbalance between the normal tissues and the lesion tissues \citep{Havaei2017, Pereira2016, Randhawa2016}.

Such approaches, however, are generally memory and computation consuming, as the surrounding area of each pixel should be included to provide contextual information in classifying each pixel. Inspired by the pioneering work of Long et al.\citep{Long2015} where a fully convolutional network (FCN) was proposed by replacing the fully-connected layers as convolutional layers, Kamnitsas, et al proposed a 3D-Convolution-based FCN, known as DeepMedic, which won the ISLES 2015 and BraTS 2015 challenges \citep{Kamnitsas2016, Kamnitsas2017}. In ISLES 2015 dataset, it is able to detect sub-acute ischemic stroke lesion of 34 out of 36 patients, and achieves a Dice coefficient of 0.59 on the test dataset \citep{Kamnitsas2017}. In BraTS 2015 challenge, it achieves a Dice coefficient of 0.85 in segmenting the tumor tissues\citep{Kamnitsas2016}. The promising results in ISLES and BraTS shows the great potential of deep learning in the brain tissue segmentation tasks. However, the MRI images provided in the challenge datasets are acquired for scientific usage with a high resolution of $1\times 1 \times 1$mm per voxel. In clinical image, the image slices are usually much thicker, typically $5$mm, and the data cannot be preprocessed with methods such as brain extraction, cerebrospinal fluid removal, standardization. Therefore, the DeepMedic developed for BraTS and ISLES datasets cannot be applied directly in clinical data.

Recently, clinical diffusion weighted image (DWI) data has been applied to lesion segmentation of acute ischemic stroke base on deep learning technique and presents a very promising results\citep{Chen2017}, where a network which combines two Deconvolution Network (DeconvNet) \citep{Noh2015} is developed and trained on a clinical dataset of 741 acute ischemic stroke patients. A multi-scale CNN is further developed to remove potential false positives. The mean dice coefficient, mean number of false positives and the mean number of false negatives achieved in \citep{Chen2017} are 0.67, 3.27 and 4.07, respectively. While the former two are reasonable, the latter is relatively too large. In clinical diagnosis, the false negatives (FNs) are more fatal compared to false positives (FPs) due to the fact that the FPs can be possibly filtered by doctors while the FNs, which are sometimes too subtle to be noticed, will lead to severe misdiagnosis accident when the misclassified lesion is the only lesion in the brain. This motivates us to study how to further reduce the FNs to develop a practical deep-learning based stroke lesion segmentation method for clinical diagnosis usage.

In stroke diagnosis, only DWI is not sufficient for the diagnosis of ischemic stroke due to the T2 shine-through effects and chemical shift artifact. Therefore, more acquisition parameters, such as Apparent Diffusion Coefficient (ADC) and T2-weighted images (T2WI), should be included in the diagnosis \citep{Kaesemann2014}. In this paper, we will show that the lesions can be preciously detected by using DWI, ADC and T2WI in the diagnosis. In the clinical work, we catch sight of acute and subacute strokes frequently coexist in the brain tissues, making them difficult to be clearly separated in segmentation. Moreover, subacute stroke lesions also have its clinical significance, and are usually required to be evaluated in clinical diagnosis. Therefore, we propose to segment both acute and subacute ischemic stroke lesions to provide useful reference for clinicians.

For improving the performance of a CNN, it is natural to use more convolution layers, such that more features can be extracted. However, the performance will become worse if we stack too many convolution layers, as the initial convolution layers cannot be trained due to the gradient vanishing problem. Recently, a so-called residual network (ResNet) is proposed to make the network much deeper. Instead of simply stack more convolution layers, many skip connections are added between layers of the network. Such structure allows the gradient to pass backward through the skip connections, and all the convolution layers are able to be updated to extract features since the first training epoch. In this paper, we propose a residual-structured fully convolutional network (Res-FCN) for brain ischemic lesion segmentation. Specifically, we collected the clinical data of 212 ischemic stroke patients from Nankai University affiliated Tianjin Huanhu Hospital, where 115 of them are used for training, and 97 of them for testing. By using multi-modal MRI images, i.e., DWI, ADC and T2WI, the proposed Res-FCN is able to achieve a mean number of false negatives of 1.515 per patient. The result achieved in our work sheds a light on the use of automated segmentation in clinical diagnosis.

\begin{figure}
  \centering
  \includegraphics[width=0.8\textwidth]{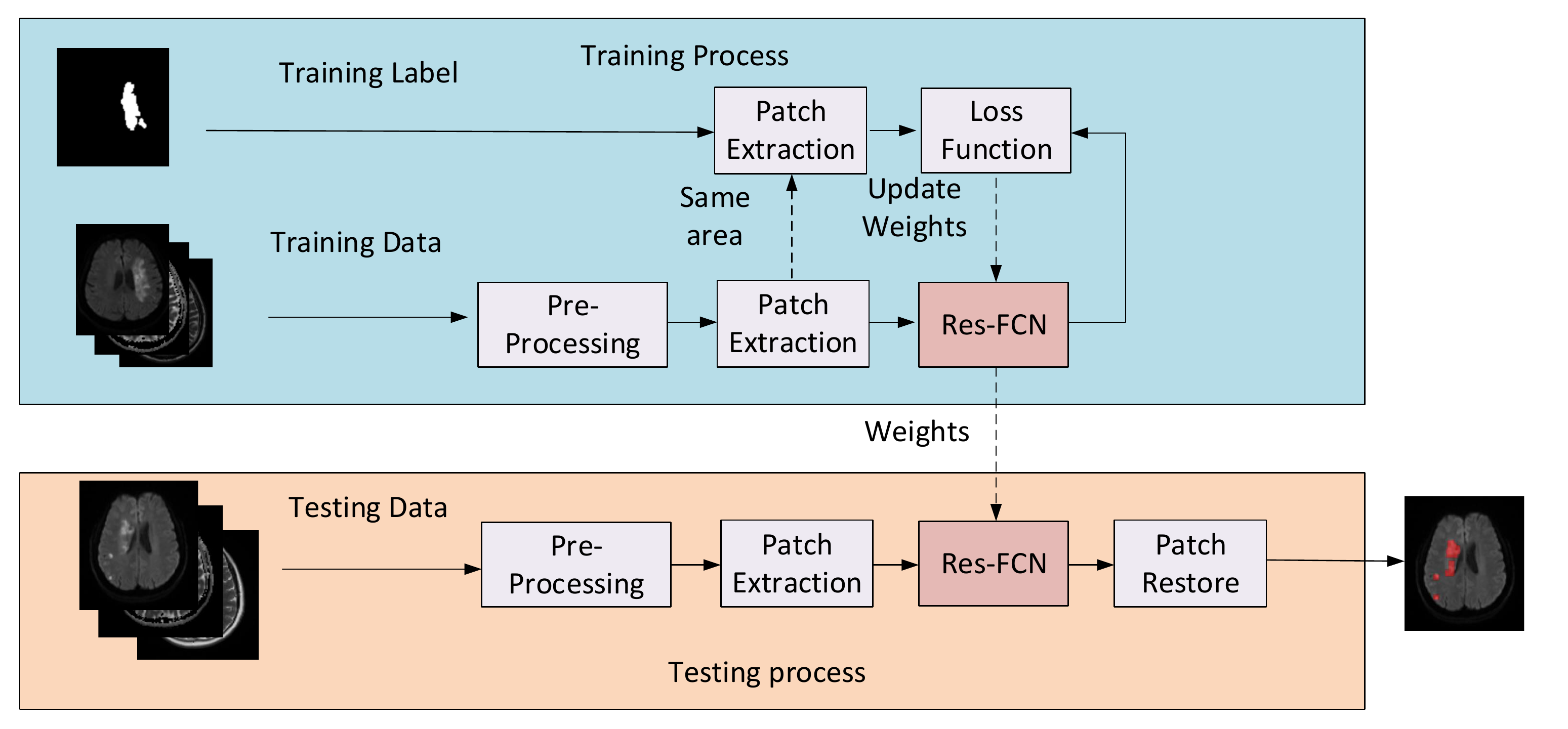}
  \caption{The whole pipeline of ischemic stroke lesion segmentation.}\label{fig_wholeprocess}
\end{figure}

\section{Method}

Fig. \ref{fig_wholeprocess} shows the whole pipeline of our proposed residual-structured fully convolutional network (Res-FCN). The DWI, ADC and T2WI images are jointly used as input, and the output is a binary segmentation of stroke lesions. The Res-FCN is first trained by using the images with manually annotated lesions to extract features. In the testing process, the images that unknown to Res-FCN is used to generate predictions.

Before giving more detailed descriptions, here we briefly introduce the outline. Specifically, in the training process, images are normalized to zero mean and unit variation in pre-processing, and patches with the size of $N^2$ is then extracted. The patches are processed by the Res-FCN, and the results are evaluated by a loss function according to the predicted results and the ground truth of the corresponding area. The weights in the Res-FCN is updated by using a gradient-based method to minimize the loss function that measures the distance between the predicted segmentation and the ground truth. The detail of the loss function will be introduced later in this section.

In the testing process, the images are cropped to patches with size of $N^2$ after normalization, and the Res-FCN with the trained weights are used for prediction. As the output for the $i$-th pixel $\hat{x}_i\in[0,1]$ which indicates the probability for the $i$-th pixel to be classified as lesion tissue, a threshold $\delta$ is used to generate the final binary segmentation, where the final binary output $\tilde{x}_i=1$ if $\hat{x}_i\geq \delta$, and $\tilde{x}_i=0$ otherwise. The predicted binary segmentation label patches are finally restored to its corresponding position to generate the final output. The details of our proposed method will be introduced in detail in the following subsections.

\subsection{Patch Extraction and Data Augmentation}

Note that the ischemic stroke lesions are typically small compared the normal brain tissues, which further reduces the size of training dataset if we use the data in slice level. Moreover, as the lesion segmentation is performed in pixel level, using the whole slice as input will generate significant class imbalance in data. Therefore, we extract 2D image patches of size $N^2$ as input.

In particular, we extract the patches by using a sliding window scheme with the window size $N^2$, and a sliding step of $N/8$ for each step. To balance the number of the normal and lesion pixels, we only include the patches with lesions into the training dataset. With a relatively large $N$, a significant part of the normal tissue will also be included, and their features can still be learnt by the network.

As each patient has a limited number of lesions, there is only a small number of patches available after patch extraction. It is thus necessary to generate a large number of patches to tune the massive network parameters. Therefore, we perform data augmentation by horizontally flipping and randomly rotating the extracted patches. Moreover, our patch extraction scheme can also be regarded as data augmentation by oversampling the pixels with lesions.

\subsection{Network Architecture}
\begin{figure}
\begin{center}
		\includegraphics[width=.6\textwidth]{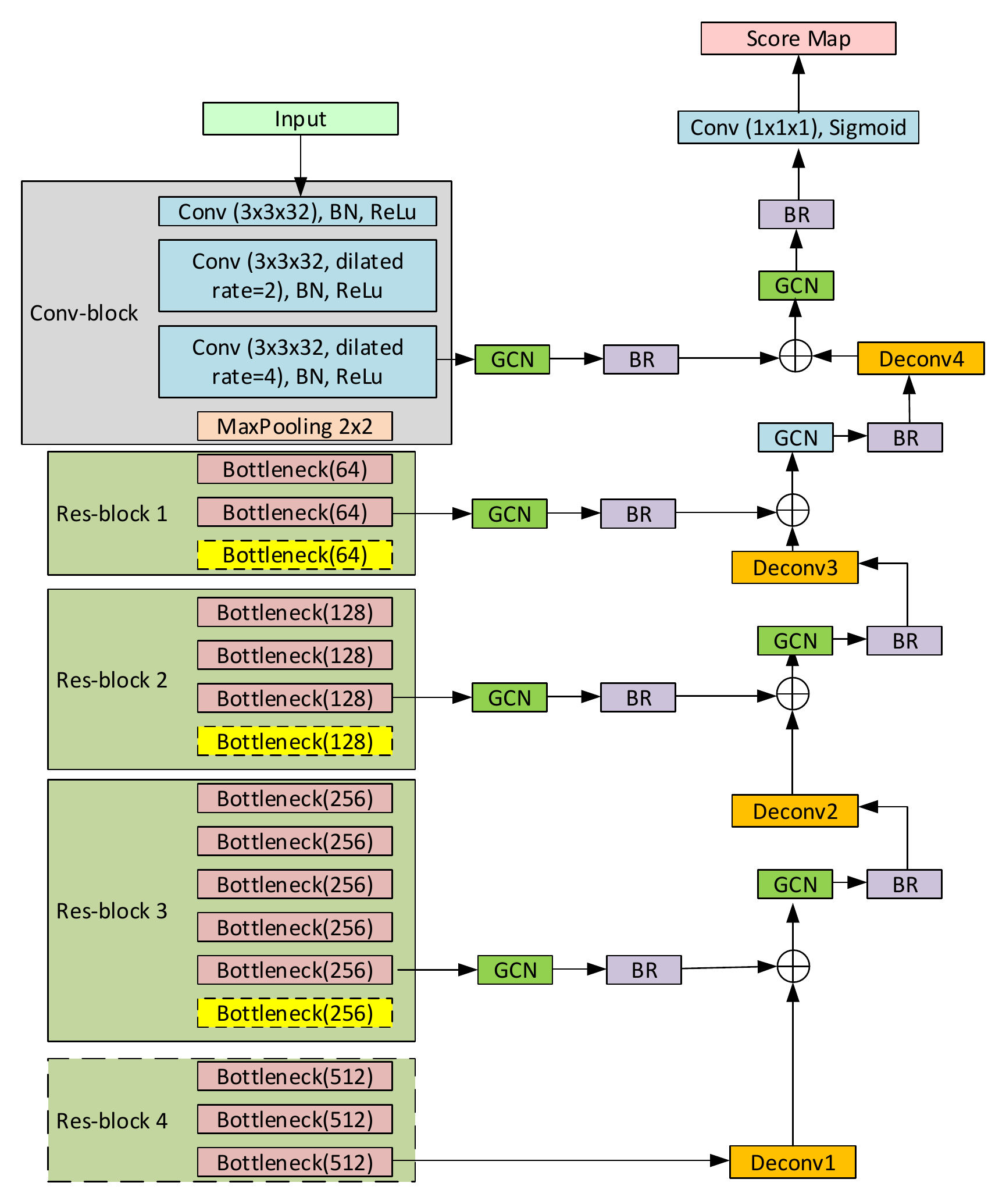}
		\caption{The architecture of the proposed network. The bottleneck block with downsampling, i.e., with convolution layers with stride 2, is highlighted in yellow. GCN: Global Convolutional Network. BR: Boundary Refinement. Conv: Convolutional layer. Deconv: Deconvolutional layer.}\label{fig_network}
\end{center}
\end{figure}

Fig. \ref{fig_network} shows the network structure of our proposed method. In this paper, we propose to use the residual network (ResNet) \citep{He2016} as the base network for feature extraction, and deconvolution with stride 2 for score map reconstruction. Several shortcuts with global convolutional network (GCN) blocks and boundary refinement (BR) blocks  \citep{Peng2017}  are used to provide higher level feature for reconstruction.

The details of the bottleneck block, GCN block, and BR block are shown in Fig. \ref{fig_block}.

\subsubsection{Convolutional layers}

\begin{figure}
  \centering
  \includegraphics[width=\textwidth]{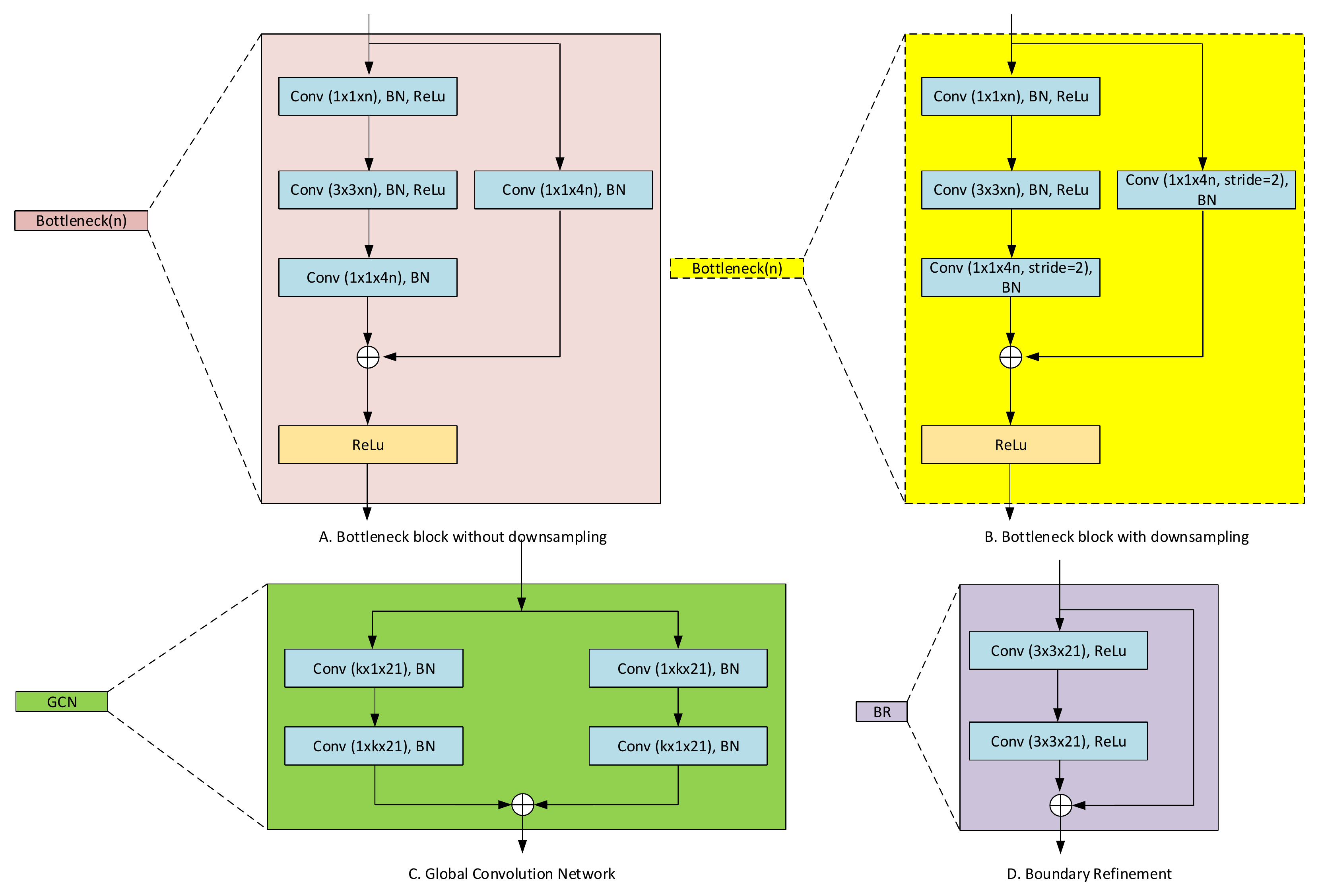}
  \caption{Illustration the details of (A) bottleneck block without downsampling, (B) bottleneck block with downsampling, (C) global convolution network (GCN) block, and (D) boundary refinement (BR) block. }\label{fig_block}
\end{figure}
The basic building block to construct a CNN is the convolutional layer. Several layers can be stacked on top of each other. Each convolutional layer can be understood as extracting features from the preceding layer, and produce the feature maps as output.

Each feature map $\mathbf{O}_s$ is associated with one kernel. The feature map $\mathbf{O}_s$ is computed as
\begin{equation}\label{conv}
	\mathbf{O}_s=\sum_{r}\mathbf{W}_{s}*\mathbf{X}_r+b_s,
\end{equation}
where $\mathbf{W}_s$ is the kernel and $b_s$ is the bias term. $*$ denotes convolution operator. $\mathbf{X}_r$ is the $r$-th channel of the input. For instance, for the first convolutional layer, the input is the stack of DWI, ADC and T2WI image patches with size $N\times N \times 3$, and $\mathbf{X}_r$ represents the $r$-th channel of the original MRI image, for $r=1,2,3$. For the subsequent convolutional layers whose input is  a $M\times M\times R$. In this case $\mathbf{X}_r$ is the output of the $r$-th feature map of the preceding layer whose size is $M\times M$, for $r=1,2,\cdots, R$, where $R$ is the number of feature maps of the preceding convolution layer.

During the training process, each convolutional layer is able to learn the features at different levels via the gradient-based method, such as the stochastic gradient descent, on a dedicated loss function related to the misclassification error, and the gradient for each layer can be computed by using the back-propagation (BP) algorithm\citep{LeCun2015}.

One of the drawback of convolutional layers is that each kernel size (typically $3\times 3$) has a limited receptive field, making it difficult to extract features from sufficiently large scale of contextual information in one layer. If we use larger kernels, the number of parameters to be tuned will grow exponentially, leading to prohibitively high computational and memory cost. More importantly, a much larger number of training data will be required if we have more parameters, which is impossible in biomedical image segmentation due to the limited number of data. Therefore, we propose to use two methods to compensate such shortcoming.

\begin{figure}
  \centering
  \includegraphics[width=0.6\textwidth]{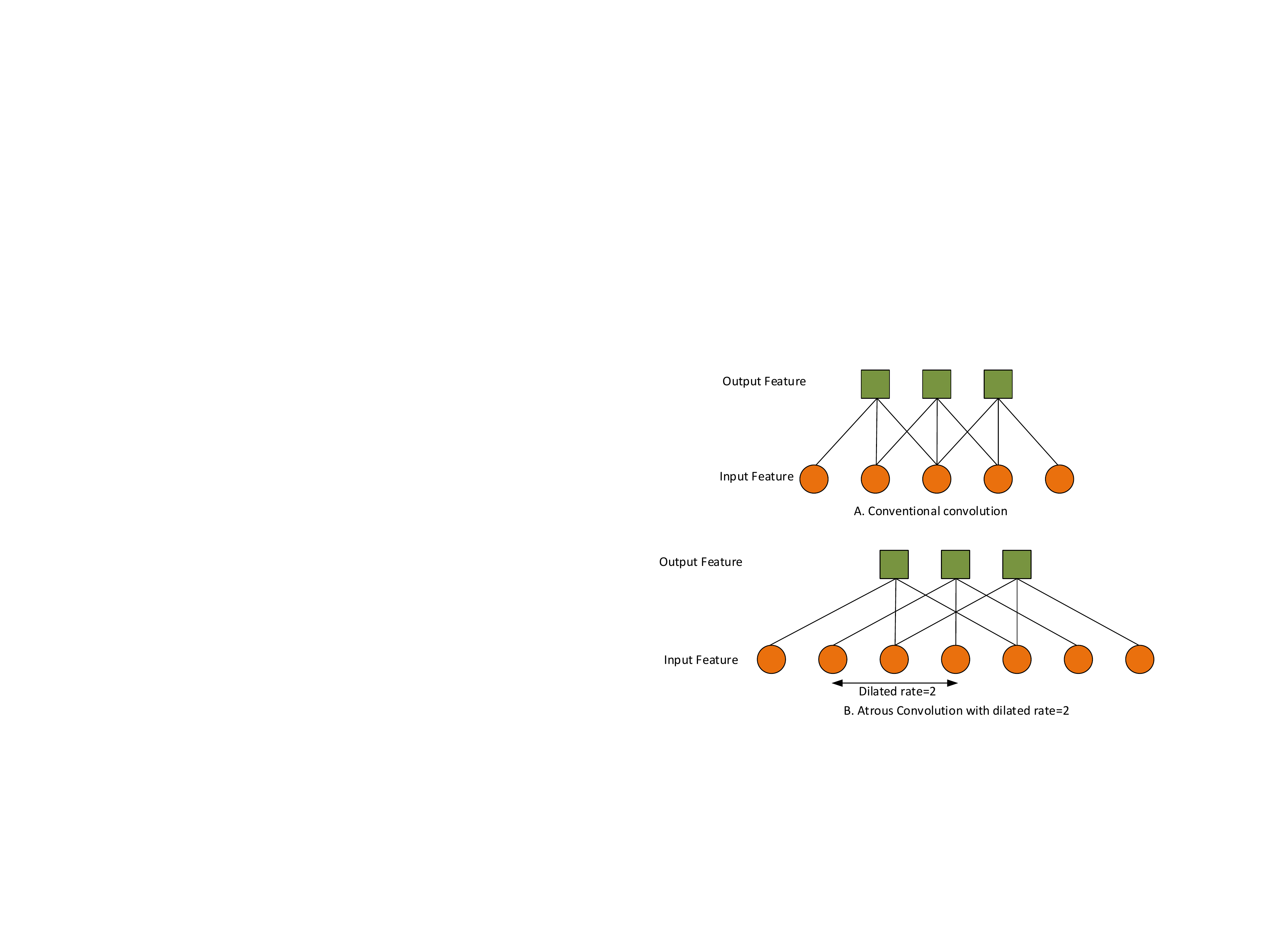}
  \caption{Graphic illustration of convolution operation. (A) Conventional convolution. (B) Atrous convolution with dilated rate $d=2$.}\label{fig_conv}
\end{figure}

First, inspired by DeepLab method \citep{Chen2016}, we propose to use a stack of atrous convolution (also known as dilated convolution) to extract multi-scale features from the inputs. The atrous convolution is operated as
\begin{equation}
	O[i,j]=\sum_{s=-\lceil S/2\rceil}^{\lfloor S/2 \rfloor} \sum_{t=-\lceil T/2\rceil}^{\lfloor T/2\rfloor} X_r[i+d\cdot s,j+d\cdot t] W[s,t],
\end{equation}
where $O[i,j]$, $X[i,j]$ and $W[i,j]$ denote the $(i,j)$-th entry in the output feature map, input image, and the $r$-th convolutional kernel, respectively. $d$ is the dilated rate. With $d=1$, the dilated convolution converges to the conventional convolution operation in (\ref{conv}). Fig. \ref{fig_conv} plots an visualized example of dilated convolution. As we can see, the atrous convolution can be interpreted as conventional convolution with a ``hole'' of size $d-1$, which enlarges the receptive area of a kernel without increasing the kernel size and the number of parameters. In our work, we propose to adopt a stack of atrous convolution layers with rate $1,2$ and 4 to extract multi-scale features from the original images.

Second, we adopt the GCN and BR blocks proposed in \citep{Peng2017} to extract more contextual information. Instead of directly using larger kernel, the GCN block employs a combination of $1\times k+k\times 1$ and $k\times 1+1\times k$ subsequently to extract the features with receptive field of $k\times k$. Compared to the $k\times k$ kernel, the GCN structure reduces the number of parameters by a factor of $\frac{2}{k}$, which enables the probability to extract more features with considerably small number of parameters. \citep{Peng2017} justifies that the GCN structure shown in Fig. \ref{fig_block}C is able to achieve better performance over that with a $k\times k$ kernel and that with a stack of small size kernels.

\subsubsection{Pooling}

It combines spatially nearby features in the feature map. Such combination makes the representation more compact and invariant to small image changes. Moreover, it also reduces the computational load for subsequent layers. In our work, we use max-pooling after the first stack of convolutional layers, and use convolution operation with stride 2 in the bottleneck blocks shown in Fig. \ref{fig_block}B.

\subsubsection{Bottleneck Block}

The bottleneck block is the basic block in a ResNet with more than 34 convolutional layers\footnote{In ResNet-34 or ResNet-18, a simple block is used\citep{He2016}. In our method, we propose to build the network based on ResNet-50, and therefore a bottleneck block is used in this paper.}, which are depicted in Fig. \ref{fig_block}A and Fig. \ref{fig_block}B. Specifically, we first use a $1\times 1$ convolutional layer with $n$ kernels as a bottleneck layer to reduce dimensionality of the features. Then we use a $3\times 3$ convolutional layer with $n$ kernels, and finally another $1\times 1$ convolutional layer with $4n$ kernels to restore the depth. A skip connection with a $1\times 1$ convolution with $4n$ kernels is used so that the input and output have the same size. Finally, the input of the block and the output of the final convolutional layer are added, and an activation is used after the summation. The number of filters are doubled every residual block, and at the same time their height and width are halved by using a convolutional layer with stride 2.

As shown in Fig. \ref{fig_block}A, instead of learning from the whole image, the convolutional layer of each bottleneck block learns the residual part $h(\mathbf{X})=f(\mathbf{X})-\mathbf{X}$, where $\mathbf{X}$ and $f(\mathbf{X})$ denote the input and output feature maps, respectively. Thanks to the skip connection, the ResNet is able to update the deepest convolutional layers, i.e., the first convolution block shown in Fig. \ref{fig_network}, from the beginning, and solves gradient vanishing problem in a dedicated way. The bottleneck structure enables us to build deeper network and extract more features, which makes the segmentations and classifications more accurate.
\subsubsection{Activation}
It is used to introduce non-linearity to each artificial neuron. In our method, a rectangled linear unit (ReLu) is used as an activation function in all layers except the last convolution layer. The ReLu function is defined as
\begin{equation}\label{relu}
  f(x)=\max(0,x).
\end{equation}
The sigmoid function is used in the last convolution layer to construct a probability map of segmentation, which is defined as
\begin{equation}\label{sigmoid}
  f(x)=\frac{1}{1-e^{-x}}.
\end{equation}

\subsubsection{Regularization}

The regularization is used to reduce over-fitting. We use $\ell_2$ norm regularization and batch normalization, such that the learning algorithm not only fits the data but also keeps the model weights as small as possible.

\subsubsection{Loss Function}

The loss function is used to measure the error. It is also the function to be minimized during training. In this paper, we propose to use negative Dice coefficient as the loss function, which is defined as
\begin{equation} \label{loss}
	f(\hat{\mathbf{x}})=-\frac{2\sum_{i\in patch}\hat{x}_i x_i+\epsilon}{\sum_{i\in patch} x_i+\sum_{i\in patch} \hat{x}_i+\epsilon},
\end{equation}
where $x_i$ and $\hat{x_i}$ denote the label of the $i$-th pixel in the ground truth and the predicted segmentation, respectively. $x_i=1$ if the $i$-th pixel is labeled as lesion, and $x_i=0$ otherwise. $\hat{x}_i \in [0,1]$ can be interpreted as the probability of the $i$-th pixel being labeled as lesion. $\epsilon>0$ is a small positive constant to avoid singularity of (\ref{loss}).

The reason we use (\ref{loss}) instead of categorical cross-entropy as loss function is that the normal and lesion pixels are significantly imbalanced even if we only extract the patches with lesions. With imbalanced classes, the network tend to category all pixels into the major class (i.e., the normal class in our work) to minimize the loss function. The loss function (\ref{loss}), however, tend to maximize the overlapping area and minimizing the non-overlapping area, such that the class imbalance problem can be mitigated.

\subsubsection{Overview of Architecture}

The proposed architecture is presented in Fig. \ref{fig_network}. In the feature extraction stage, we propose to use the ResNet 50 structure as the base network. The input patches are initially processed by a stack of three atrous convolutional layers with dilated rate 1, 2 and 4. Then four Res-blocks are used to extract more features, where the Res-blocks are composed of 3,4,6 and 3 bottleneck blocks with the filters $n=64,128,256$ and 512, respectively, as suggested in \citep{He2016}.

At the score map reconstruction stage, we use deconvolutional layers with kernel size $(3\times 3)$ and strides 2 to upsample the feature maps. Inspired by U-NET \citep{Ronneberger2015}, we propose to use the high resolution feature map as an assistance to reconstruct the score map. GCN and BR are used to extract feature maps from the summation of the upsampled score map and the feature maps from shortcut connections.

\subsection{Evaluations}

We aim to study an automated ischemic lesion segmentation method towards clinic diagnosis. The dice coefficient (DC) is a commonly used measurement in image segmentation accuracy, which is defined as
\begin{equation}\label{dc def}
  DC=\frac{2|A\bigcap B|}{|A|+|B|},
\end{equation}
where $A$ and $B$ denote the ground truth and the predicted segmentation. $|A|$ denotes the area of lesion segmentations in $A$. Moreover, we introduce the mean number of false negatives and the mean number of false positives, denoted as m\#FN and m\#FP, respectively, as the additional metric for evaluation. Such metrics evaluates the accuracy of detection in subject level, which is more important in assisting the doctors in clinical diagnosis.

\section{Experiment Results}

\subsection{Dataset and preprocessing}

In this study, 212 patients with ischemia lesions were collected from Nankai University affiliated Tianjin Huanhu Hospital, where $62\%$ of them are male, and the mean age is 56.21. All clinical images were collected from a retrospective database and anonymized prior to use. MRI measurements were acquired from three MR scanners, with two 3T MR scanners (Skyra, Siemens and Trio, Siemens) and one 1.5T MR scanner (Avanto, Siemens). T2WI were acquired using a fast spin-echo sequence. DWI were acquired using a spin-echo type echo-planar (SE-EPI) sequence two b values of 0 and 1000 s/mm$^2$. The parameters are summarized in Table \ref{tab_parameter}. Following acquisition, an ADC map was calculated using the diffusion scan raw data. The T2WI, DWI and ADC images were copy referenced to ensure same slice position so as to allow optimal image evaluation and measurement. The ischemic lesions were manually annotated by experienced experts. We use 115 of them to train and validate our network, and 97 of them for testing only.

\begin{table}
  \centering
   \caption{Parameters used in MRI acquisition.}\label{tab_parameter}
  \begin{tabular}{l|cccccccc}
  \toprule
  &\multicolumn{2}{c}{Skyra} & \multicolumn{2}{c}{Trio}& \multicolumn{2}{c}{Avanto}\\
   & DWI & T2WI &                   DWI&T2WI&                        DWI&T2WI \\
  \hline
  Repitition time (ms) & 5200 & 3950 & 3100 & 4000 & 3800 & 4410\\
  echo time (ms) & 80 & 99 & 99 & 93 & 102& 90 \\
  Flip angle ($^\circ$) & 150 & 150 & 120 & 120& 150& 150  \\
  Number of excitations & 1 & 1 & 3 & 1 & 3 & 1 \\
  Field of view ($\text{mm}^2$)& $240\times 240$ & $240\times 240$ & $200\times 200$  & $230\times 230$ & $240\times 240$ &$240\times 240$\\
  Matrix size & $130\times 130$ & $320\times 320$ &   $132\times 132$ & $320\times 320$ & $192\times 192$ & $320\times 320$\\
  Slice thickness (mm) & 5 & 5 & 6 & 6 & 5 & 5  \\
  Slice spacing (mm) & 1.5 & 1.5  & 1.8 & 1.8 & 1.5 & 1.5\\
  Number of slices & 21 & 21 & 17 & 17 & 21 & 21\\
    \bottomrule
\end{tabular}

\end{table}

We resample all images to a pixel size of $1.77\times 1.77$mm with linear interpolation, and crop the matrix to $128\times 128$. Then we register the T2WI and ADC images according to the corresponding DWI images.

The intensity of each image slice is normalized into that of zero mean and unit variance. The patch are extracted with a size $64\times 64$, and the sliding step to sample the patches is 8 pixels per step. The extracted patches are splitted into training set and validation set before data augmentation, and a proportion of 0.1 lesion patches are used for validation. The patches in the training set are augmented using the method introduced in Sec. 2, and no data augmentation is performed for validate set.
\subsection{Setup}

The hyperparameters of the proposed network are shown in Table \ref{tab_netset}. All the parameters are initialized in the way as suggested in \citep{He2016}, and the kernel regularization factor is set to be $10^{-4}$. We use the Adam method \citep{Kingma2014} with initial learning rate of $0.001$, $\beta_1=0.9$, $\beta_2=0.999$ as optimizer, and the learning rate is scaled down in a factor of $\sqrt{0.1}$ if no progress is made for 5 epoches in validation data. Without specifications, in this section, the threshold $\delta$ to generate binary segmentation is set to be $0.5$.

\begin{table*}
\caption{Network Setting} \label{tab_netset}
\begin{center}
		\begin{tabular}{ccccr}
\hline \\
  Layer & Filter size & Stride & Number of filters & Output shape \\
  \hline \\
  Input&-&-&-&$64\times 64 \times 3$ \\
  Conv-block & $3\times 3$ & 1 & 32 & $64\times 64 \times 32$\\
  Max Pooling& $2\times 2$ & 1 & -&$32\times 32\times 32$ \\
  Res-block 1 & -& - & $n=64$& $16\times 16 \times 256$\\
  Res-block 2& - & - & $n=128$& $8\times 8 \times 256$\\
  Res-block 3 & -& - & $n=256$& $4\times 4 \times 256$\\
  Res-block 4 & -& - & $n=512$& $4\times 4 \times 256$\\
  Deconv1 & $3\times 3$ & 2 & 21 & $8\times 8 \times 21$ \\
  Deconv2 & $3\times 3$ & 2 & 21 & $16\times 16 \times 21$ \\
  Deconv3 & $3\times 3$ & 2 & 21 & $32\times 32 \times 21$ \\
  Deconv4 & $3\times 3$ & 2 & 21 & $64\times 64 \times 21$ \\
  Conv & $1\times 1$ & - & 1 & $64\times 64 \times 1$ \\
  \hline
\end{tabular}
\end{center}
\end{table*}

\subsection{Implementation}

The experiments are performed on an Alienware Aurora R6 computer with an Intel Core i7-7700K CPU, 48GB RAM and Nvidia GeForce 1080Ti GPU with 11GB memory. The network is implemented on Keras with Tensorflow backend. The MR image files are stored as Neuroimaging Informatics Technology Initiative (NIfTI) format, and processed using Simple Insight Toolkit (SimpleITK). The visualized results are presented by using ITK-SNAP\citep{Yushkevich2006}.

\subsection{Results}

\begin{figure}
\begin{center}
  \includegraphics[width=0.8\textwidth]{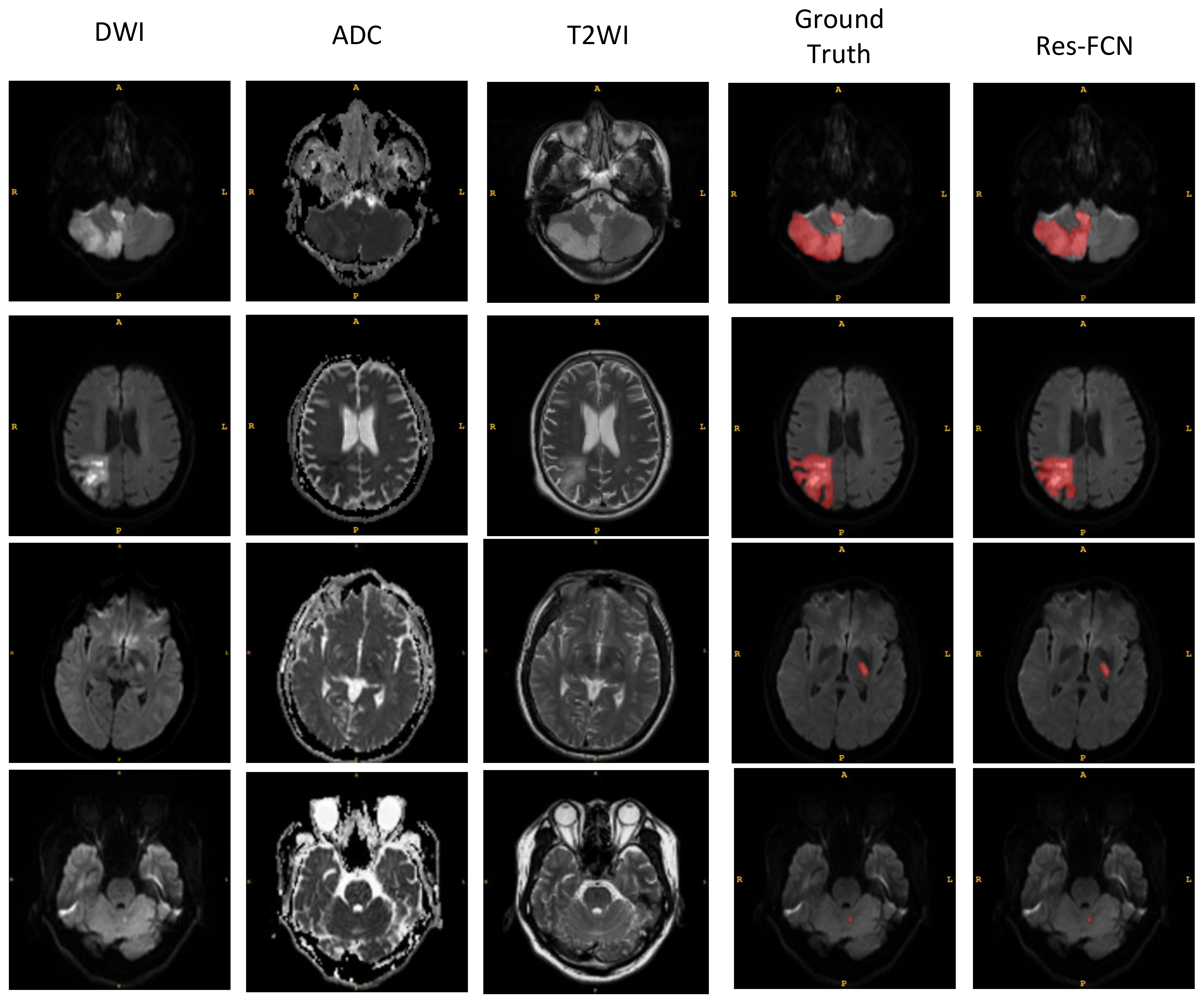}
  \caption{Examples of lesion segmentation. The first three columns show the original DWI, ADC, T2WI images, respectively. The fourth and the fifth columns show the manually annotated lesions and the segmentation results of Res-FCN, respectively. the The segmentations are depicted on the DWI, and highlighted in red. } \label{fig_example1}
\end{center}
\end{figure}

The trained network is tested on 97 patients. Fig. \ref{fig_example1} plots some examples of segmentations where the lesions are located at cerebellum, cerebral hemisphere and basal ganglia. The fourth row shows an example with a small lesion at the cerebellar vermis. As Fig. \ref{fig_example1} shows, our proposed method is sensitive to both large and small lesions. Table \ref{tab_compare1} summarizes the results on the training and testing dataset. For comparison, we also evaluate the results of U-Net \citep{Ronneberger2015} and EDD-Net \citep{Chen2017} using the same data. Note that compared to the results reported in \citep{Chen2017}, the EDD-Net achieves worse results in DC and m\#FP due to two reasons: 1) the multi-scale network (MUSCLE-Net) \citep{Chen2017} is not used to further remove false positives, and 2) the threshold to generate binary segmentation is set to be $0.5$. As we will show in the following subsection, higher DC can be achieved by adjusting the threshold $\delta$.


Despite that our network is much deeper than U-Net and EDD-Net, the Res-FCN achieves the best results on the testing dataset thanks to the residual structure of the bottleneck block which solves the gradient vanishing problem in very deep CNN, so that the convolutional layers are able to extract more features. As we can see from Table \ref{tab_compare1}, the Res-FCN is able to achieve a mean number of FNs of 1.515, which is very close to the results of a common medical image doctor.

\begin{table*}
\caption{Performance of our proposed Res-FCN with different GCN kernel size $k$, U-NET \citep{Ronneberger2015} and EDD Net \citep{Chen2017}. The bold number indicates the most significant performance.} \label{tab_compare1}
\begin{center}
		\begin{tabular}{l|cccc}
		\toprule \\
		& &U-NET & EDD Net & Res-FCN($k=9$) \\
		\hline \\
		Dice & train & 0.684 & \textbf{0.812} & 0.803 \\
		& test & 0.541 & 0.626 & \textbf{0.645}  \\
		m\#FP & train & 7.017 & 4.609& \textbf{4.504}  \\
		& test & 7.433 & 5.031 & \textbf{4.237} \\
		m\#FN & train & 0.279 & 0.139 & \textbf{0.130} \\
		& test & 1.918 &1.753  & \textbf{1.515} \\
		\bottomrule
	\end{tabular}
\end{center}
\end{table*}

%

\subsection{Discussions}

\subsubsection{Effect of GCN Kernel Size}

In our proposed network, we use the atrous convolution layer stack instead of conventional convolution as the first convolutional layers. The proposed pyramid atrous convolution shows of paramount importance in extracting multi-scale information from the original images. To illustrate this, we compare the performance with atrous convolutions and conventional convolutions on our dataset, which are summarized in Table \ref{tab_compare2}. The use of atrous convolution layer stack significantly increases the DC, and the m\#FP and m\#FN are greatly reduced, due to the fact that the atrous convolution has a larger receptive field. In particular, with atrous convolutions, the feature receptive fields are $3\times 3$, $5\times 5$ and $7\times 7$, while with conventional convolutions, the receptive field of the three convolutional layers are all $3\times 3$. With a larger receptive field, the network is able to learn with more contextual information without increasing the kernel size.

Table \ref{tab_compare2} also summarizes the results with different GCN kernel sizes, i.e., $k=5, 7$ and $9$. The DC can be significantly improved by using larger GCN kernel, which is in accordance with the observation in \citep{Peng2017}. Moreover, the mean number of FPs and FNs both decreases as the kernel size $k$ increases, indicating that more contextual information is extremely helpful in improving the performance of lesion segmentation.

\begin{table*}
\caption{Performance of our proposed Res-FCN with a stack of three atrous convolution layers with kernel sizes $3\times 3$ and dilated rate 1, 2 and 4, denoted as ``atrous'', and that with a stack of three conventional convolution layers with kernel size $3\times 3$, denoted as ``conv''. The size of GCN kernels are 5, 7 and 9.} \label{tab_compare2}
\begin{center}
		\begin{tabular}{ll|cc|cc|cc}
		\toprule \\
        &&\multicolumn{2}{c}{$k=5$}\vline &\multicolumn{2}{c}{$k=7$}\vline&\multicolumn{2}{c}{$k=9$}\\
		& & atrous&  conv &  atrous &  conv &  atrous &  conv \\
		\midrule \\
		Dice & train & 0.786 & 0.782 & 0.721 & 0.739 & 0.803&0.784\\
		& test & 0.600 & 0.589 & 0.623 & 0.568& 0.645& 0.613\\
		m\#FP & train & 3.678 & 4.521& 5.217 & 6.374 & 4.504 & 4.165 \\
		& test & 5.928 & 8.340 & 4.515 & 7.505 & 4.237& 4.536\\
		m\#FN & train & 0.113 & 0.087 & 0.243 & 0.113 & 0.130 & 0.087\\
		& test & 1.546 & 1.629 & 1.619 & 1.690 & 1.515 & 1.660\\
		\bottomrule
	\end{tabular}
\end{center}
\end{table*}

\subsubsection{Tradeoff between FNs and FPs}

Note that the output of a CNN indicates the probability that a pixel should be labeled as lesion tissue, and a threshold $\delta$ is required to convert the probability score map to a binary segmentation. The segmentation results presented previously are obtained by setting the threshold $\delta=0.5$. In fact, the choice of the threshold can be interpreted as a tradeoff between FNs and FPs. Intuitively, as the threshold $\delta$ increases, fewer pixels will be classified as stroke lesion tissues, leading to an increasing m\#FN and decreasing m\#FP.

\begin{figure}
  \centering
  \includegraphics[width=.6\textwidth]{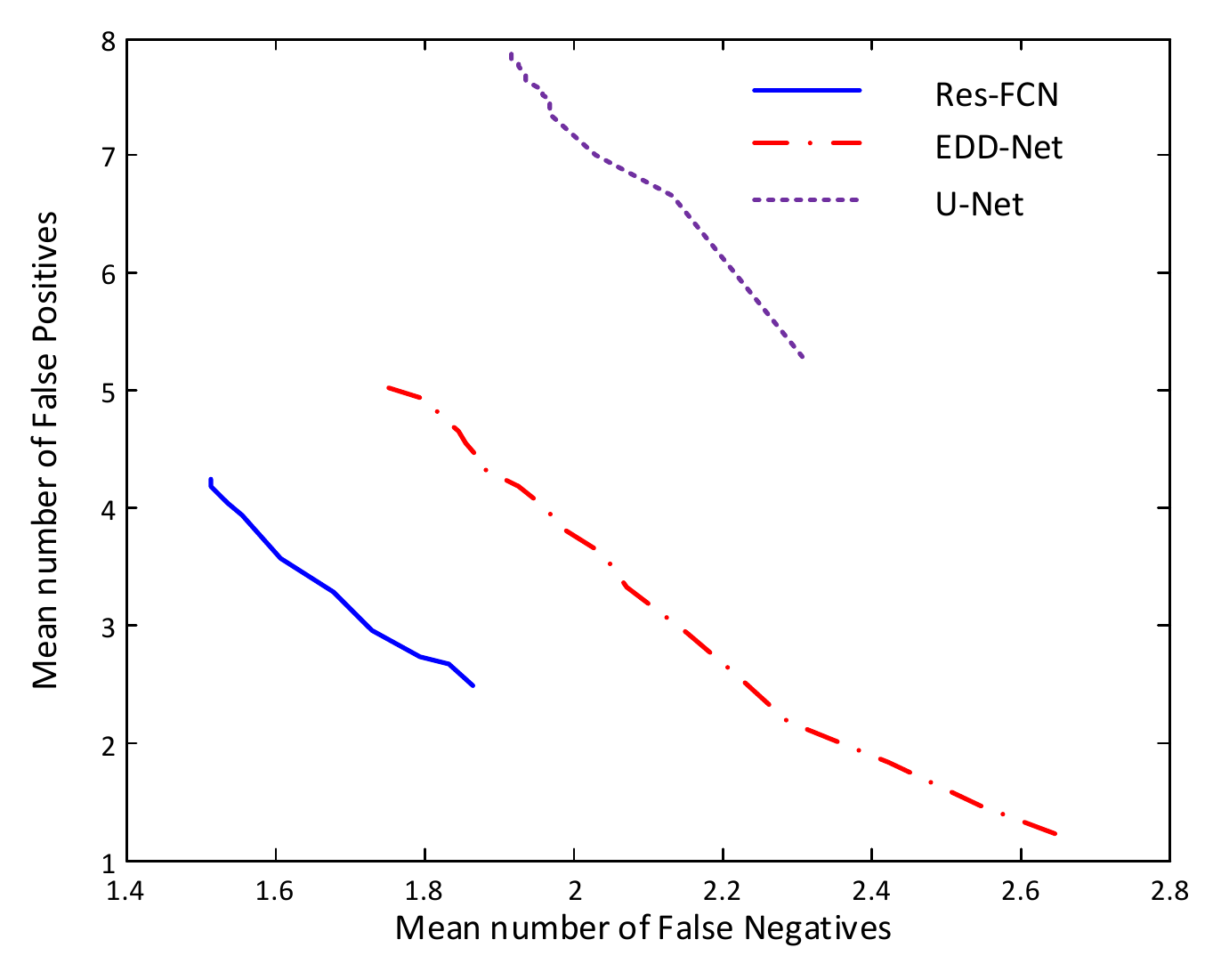}
  \caption{Mean number of false negatives versus mean number of false positives on the test dataset.}\label{fig_fnfp}
\end{figure}

Fig. \ref{fig_fnfp} plots the dependence between m\#FN and m\#FP of Res-FCN with GCN kernel size $k=9$ on the test dataset. Each dot of the curve is plotted by using different values of the threshold $\delta$, ranging from 0.5 to 1. The tradeoffs of EDD-Net and U-Net are also plotted for comparison. As the threshold $\delta$ increases, the m\#FP reduces at the expense of a higher m\#FN. As we can see from Fig. \ref{fig_fnfp}, the proposed Res-FCN presents the best FN-FP-tradeoff. In particular, for given value of m\#FN, the Res-FCN has about 2 FPs less than EDD-Net, which highlights the outstanding performance of Res-FCN in clinical diagnosis.

\begin{figure}
  \centering
  \includegraphics[width=.6\textwidth]{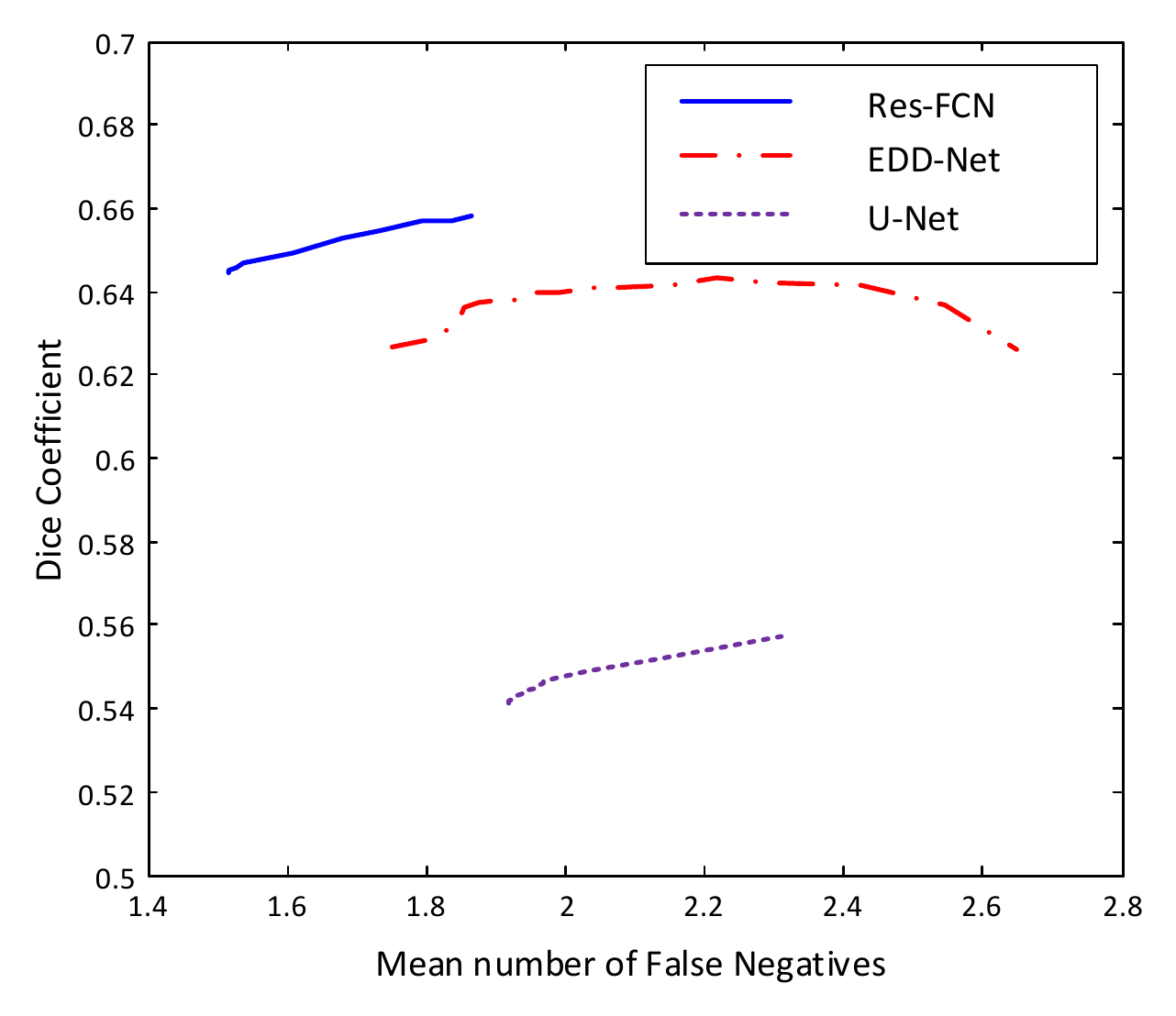}
  \caption{Mean number of false negatives versus mean dice coefficient on the test dataset.}\label{fig_fndice}
\end{figure}

Fig. \ref{fig_fndice} further plots the dependence between m\#FN and the DC with different values of the threshold $\delta$. As we can see from Fig. \ref{fig_fndice}, the Res-FCN presents the best DC performance over EDD-Net and U-Net. The U-Net and Res-FCN presents a monotonically increase with the threshold $\delta$. With $\delta=1$, the Res-FCN achieves a DC of $0.658$ at the expense that m\#FN increases to $1.866$. With EDD-Net, the DC is maximized at $\delta=0.9$ according to our experiment, where the DC is $0.644$ and the m\#FN increases to $2.217$. Compared to EDD-Net, our proposed Res-FCN is able to achieve higher segmentation accuracy with much fewer misclassifications.

\begin{table*}
\caption{Performance of Res-FCN, EDD-Net and U-Net on the testing dataset with the threshold $\delta=1$ for U-Net and Res-FCN, and $\delta=0.9$ for EDD-Net. } \label{tab_compare3}
\begin{center}
		\begin{tabular}{l|ccccccc}
		\toprule \\
		& U-Net& EDD-Net & Res-FCN($k=9$)  \\
		\midrule \\
        Dice & 0.557 & 0.644 & 0.658 \\
        m\#FP & 5.289 & 2.588 & 2.485 \\
        m\#FN & 2.309 & 2.217 & 1.866\\
		\bottomrule
	\end{tabular}
\end{center}
\end{table*}

Table \ref{tab_compare3} summarizes the performance on the testing dataset where the threshold $\delta$ is set to be the value that maximizes the DC. As we can see from Table \ref{tab_compare3}, our proposed Res-FCN presents the highest DC and the lowest misdiagnosis.  Compared to the results presented in \citep{Chen2017}, much lower number of false negatives and false positives is presented by using EDD-Net on our clinical dataset, due to the fact that we use multi-modal images, i.e., DWI, ADC and T2WI, instead of single-modal images, i.e., DWI, as input. Through the mutual corroboration of the different sequence, the detection of the stroke is more accurate, and the artifact can be avoided effectively. It highlights the importance to include multi-modal MRI images in designing stroke lesion segmentation algorithms.

\section{Conclusion}

In this paper, we presented a fully automated ischemia lesion segmentation method based on fully convolutional neural network. The DWI, ADC and T2WI images are used as input, and a very deep CNN is built and efficiently trained thanks to the residual structure. The proposed Res-FCN network presents a high segmentation accuracy on the clinical MRI image with a dice coefficient of 0.645. More importantly, it presents very low false negatives, with a mean number of 1.515 per patient, which is of paramount importance in avoiding misdiagnosis in clinical scenario. The false negatives can reach a value that is very close to the results of a common medical image doctor, making the method exceptive for a real clinical application.

Note that there are several limitations in the present study. First, multi-modal images have been demonstrated to provide helpful information for determining stroke ages in ischemic stroke. Although our data include acute and subacute ischemic stroke lesions, we have not made a distinction between them. Second, our research did not include detection of hemorrhagic transformation, which is a complication of ischemic stroke. MRI-derived information about localization, timing and pathophysiology could improve decisions regarding acute management and secondary prevention. In the future, more MR images should be collected, including all types of stroke. The classification of stroke should be perfected, in order to guide the clinical treatment.

\section*{Funding}
This work is supported by the 2016 Grants-in-Aid for Scientific Research, Ministry of Education,
Culture, Sports,
Science and Technology, Japan [project number 16K00335], the National Natural Science Foundation of China [grant numbers 61501262, 61571244], by Tianjin Research Program of Application Foundation and Advanced Technology [grant numbers 15JCYBJC51600 and 16YFZCSF00540].

\section*{\refname}
\bibliographystyle{plainnat}
\bibliography{ref}

\begin{thebibliography}{28}
\providecommand{\natexlab}[1]{#1}
\providecommand{\url}[1]{\texttt{#1}}
\expandafter\ifx\csname urlstyle\endcsname\relax
  \providecommand{\doi}[1]{doi: #1}\else
  \providecommand{\doi}{doi: \begingroup \urlstyle{rm}\Url}\fi

\bibitem[Ashton et~al.(2003)Ashton, Takahashi, Berg, Goodman, Totterman, and
  Ekholm]{Ashton2003}
Edward~A. Ashton, Chihiro Takahashi, Michel~J. Berg, Andrew Goodman, Saara
  Totterman, and Sven Ekholm.
\newblock Accuracy and reproducibility of manual and semiautomated
  quantification of ms lesions by mri.
\newblock \emph{Journal of Magnetic Resonance Imaging}, 17\penalty0
  (3):\penalty0 300--308, 2003.
\newblock ISSN 1522-2586.
\newblock \doi{10.1002/jmri.10258}.

\bibitem[Chen et~al.(2017)Chen, Bentley, and Rueckert]{Chen2017}
Liang Chen, Paul Bentley, and Daniel Rueckert.
\newblock Fully automatic acute ischemic lesion segmentation in {DWI} using
  convolutional neural networks.
\newblock \emph{NeuroImage: Clinical}, 15:\penalty0 633--643, January 2017.
\newblock ISSN 2213-1582.
\newblock \doi{10.1016/j.nicl.2017.06.016}.

\bibitem[Chen et~al.(2016)Chen, Papandreou, Kokkinos, Murphy, and
  Yuille]{Chen2016}
Liang{-}Chieh Chen, George Papandreou, Iasonas Kokkinos, Kevin Murphy, and
  Alan~L. Yuille.
\newblock Deeplab: Semantic image segmentation with deep convolutional nets,
  atrous convolution, and fully connected crfs.
\newblock \emph{CoRR}, abs/1606.00915, 2016.

\bibitem[Crichton et~al.(2016)Crichton, Bray, McKevitt, Rudd, and
  Wolfe]{Crichton2016}
Siobhan~L Crichton, Benjamin~D Bray, Christopher McKevitt, Anthony~G Rudd, and
  Charles D~A Wolfe.
\newblock Patient outcomes up to 15 years after stroke: survival, disability,
  quality of life, cognition and mental health.
\newblock \emph{Journal of Neurology, Neurosurgery \& Psychiatry}, 87\penalty0
  (10):\penalty0 1091--1098, 2016.
\newblock ISSN 0022-3050.
\newblock \doi{10.1136/jnnp-2016-313361}.

\bibitem[Doyle et~al.(2013)Doyle, Vasseur, Dojat, and Forbes]{Doyle2013}
S.~Doyle, F.~Vasseur, M.~Dojat, and F.~Forbes.
\newblock Fully automatic brain tumor segmentation from multiple mr sequences
  using hidden markov fields and variational em.
\newblock \emph{Proceedings of NCI-MICCAI BRATS.}, 1:\penalty0 18--22, 2013.

\bibitem[Ellwaa et~al.(2016)Ellwaa, Hussein, AlNaggar, Zidan, Zaki, Ismail, and
  Ghanem]{Ellwaa2016}
Abdelrahman Ellwaa, Ahmed Hussein, Essam AlNaggar, Mahmoud Zidan, Michael Zaki,
  Mohamed~A. Ismail, and Nagia~M. Ghanem.
\newblock Brain tumor segmantation using random forest trained on iteratively
  selected patients.
\newblock In Alessandro Crimi, Bjoern Menze, Oskar Maier, Mauricio Reyes,
  Stefan Winzeck, and Heinz Handels, editors, \emph{Brainlesion: Glioma,
  Multiple Sclerosis, Stroke and Traumatic Brain Injuries}, pages 129--137,
  Cham, 2016. Springer International Publishing.
\newblock ISBN 978-3-319-55524-9.

\bibitem[Gooya et~al.(2011)Gooya, Pohl, Bilello, Biros, and
  Davatzikos]{Gooya2011}
Ali Gooya, Kilian~M. Pohl, Michel Bilello, George Biros, and Christos
  Davatzikos.
\newblock \emph{Joint Segmentation and Deformable Registration of Brain Scans
  Guided by a Tumor Growth Model}, pages 532--540.
\newblock Springer Berlin Heidelberg, Berlin, Heidelberg, 2011.
\newblock ISBN 978-3-642-23629-7.

\bibitem[Havaei et~al.(2017)Havaei, Davy, Warde-Farley, Biard, Courville,
  Bengio, Pal, Jodoin, and Larochelle]{Havaei2017}
Mohammad Havaei, Axel Davy, David Warde-Farley, Antoine Biard, Aaron Courville,
  Yoshua Bengio, Chris Pal, Pierre-Marc Jodoin, and Hugo Larochelle.
\newblock Brain tumor segmentation with {Deep} {Neural} {Networks}.
\newblock \emph{Medical Image Analysis}, 35:\penalty0 18--31, January 2017.
\newblock ISSN 1361-8415.
\newblock \doi{10.1016/j.media.2016.05.004}.

\bibitem[He et~al.(2016)He, Zhang, Ren, and Sun]{He2016}
K.~He, X.~Zhang, S.~Ren, and J.~Sun.
\newblock Deep residual learning for image recognition.
\newblock In \emph{2016 IEEE Conference on Computer Vision and Pattern
  Recognition (CVPR)}, pages 770--778, June 2016.
\newblock \doi{10.1109/CVPR.2016.90}.

\bibitem[Kaesemann et~al.(2014)Kaesemann, Thomalla, Cheng, Treszl, Fiehler, and
  Forkert]{Kaesemann2014}
Philipp Kaesemann, GÃ¶tz Thomalla, Bastian Cheng, Andras Treszl, Jens
  Fiehler, and Nils~Daniel Forkert.
\newblock Impact of severe extracranial ica stenosis on mri perfusion and
  diffusion parameters in acute ischemic stroke.
\newblock \emph{Frontiers in Neurology}, 5:\penalty0 254, 2014.
\newblock ISSN 1664-2295.
\newblock \doi{10.3389/fneur.2014.00254}.

\bibitem[Kamnitsas et~al.(2016)Kamnitsas, Ferrante, Parisot, Ledig, Nori,
  Criminisi, Rueckert, and Glocker]{Kamnitsas2016}
Konstantinos Kamnitsas, Enzo Ferrante, Sarah Parisot, Christian Ledig,
  Aditya~V. Nori, Antonio Criminisi, Daniel Rueckert, and Ben Glocker.
\newblock Deepmedic for brain tumor segmentation.
\newblock In Alessandro Crimi, Bjoern Menze, Oskar Maier, Mauricio Reyes,
  Stefan Winzeck, and Heinz Handels, editors, \emph{Brainlesion: Glioma,
  Multiple Sclerosis, Stroke and Traumatic Brain Injuries}, pages 138--149,
  Cham, 2016. Springer International Publishing.
\newblock ISBN 978-3-319-55524-9.

\bibitem[Kamnitsas et~al.(2017)Kamnitsas, Ledig, Newcombe, Simpson, Kane,
  Menon, Rueckert, and Glocker]{Kamnitsas2017}
Konstantinos Kamnitsas, Christian Ledig, Virginia F.~J. Newcombe, Joanna~P.
  Simpson, Andrew~D. Kane, David~K. Menon, Daniel Rueckert, and Ben Glocker.
\newblock Efficient multi-scale 3d {CNN} with fully connected {CRF} for
  accurate brain lesion segmentation.
\newblock \emph{Medical Image Analysis}, 36:\penalty0 61--78, February 2017.
\newblock ISSN 1361-8415.
\newblock \doi{10.1016/j.media.2016.10.004}.

\bibitem[Kingma and Ba(2014)]{Kingma2014}
Diederik~P. Kingma and Jimmy Ba.
\newblock Adam: {A} method for stochastic optimization.
\newblock \emph{CoRR}, abs/1412.6980, 2014.

\bibitem[Le~Folgoc et~al.(2016)Le~Folgoc, Nori, Ancha, and
  Criminisi]{LeFolgoc2016}
Loic Le~Folgoc, Aditya~V. Nori, Siddharth Ancha, and Antonio Criminisi.
\newblock Lifted auto-context forests for brain tumour segmentation.
\newblock In Alessandro Crimi, Bjoern Menze, Oskar Maier, Mauricio Reyes,
  Stefan Winzeck, and Heinz Handels, editors, \emph{Brainlesion: Glioma,
  Multiple Sclerosis, Stroke and Traumatic Brain Injuries}, pages 171--183,
  Cham, 2016. Springer International Publishing.
\newblock ISBN 978-3-319-55524-9.

\bibitem[LeCun et~al.(2015)LeCun, Bengio, and Hinton]{LeCun2015}
Yann LeCun, Yoshua Bengio, and Geoffrey Hinton.
\newblock Deep learning.
\newblock \emph{Nature}, 521:\penalty0 436--444, May 2015.

\bibitem[Lefkovits et~al.(2016)Lefkovits, Lefkovits, and
  Szil{\'a}gyi]{Lefkovits2016}
L{\'a}szl{\'o} Lefkovits, Szid{\'o}nia Lefkovits, and L{\'a}szl{\'o}
  Szil{\'a}gyi.
\newblock Brain tumor segmentation with optimized random forest.
\newblock In Alessandro Crimi, Bjoern Menze, Oskar Maier, Mauricio Reyes,
  Stefan Winzeck, and Heinz Handels, editors, \emph{Brainlesion: Glioma,
  Multiple Sclerosis, Stroke and Traumatic Brain Injuries}, pages 88--99, Cham,
  2016. Springer International Publishing.
\newblock ISBN 978-3-319-55524-9.

\bibitem[Liu et~al.(2014)Liu, Niethammer, Kwitt, McCormick, and
  Aylward]{Liu2014}
Xiaoxiao Liu, Marc Niethammer, Roland Kwitt, Matthew McCormick, and Stephen
  Aylward.
\newblock \emph{Low-Rank to the Rescue -- Atlas-Based Analyses in the Presence
  of Pathologies}, pages 97--104.
\newblock Springer International Publishing, Cham, 2014.
\newblock ISBN 978-3-319-10443-0.

\bibitem[Long et~al.(2015)Long, Shelhamer, and Darrell]{Long2015}
J.~Long, E.~Shelhamer, and T.~Darrell.
\newblock Fully convolutional networks for semantic segmentation.
\newblock In \emph{2015 IEEE Conference on Computer Vision and Pattern
  Recognition (CVPR)}, pages 3431--3440, June 2015.
\newblock \doi{10.1109/CVPR.2015.7298965}.

\bibitem[Maier et~al.(2017)Maier, Menze, von~der Gablentz, Häni, Heinrich,
  Liebrand, Winzeck, Basit, Bentley, Chen, Christiaens, Dutil, Egger, Feng,
  Glocker, Götz, Haeck, Halme, Havaei, Iftekharuddin, Jodoin, Kamnitsas,
  Kellner, Korvenoja, Larochelle, Ledig, Lee, Maes, Mahmood, Maier-Hein,
  McKinley, Muschelli, Pal, Pei, Rangarajan, Reza, Robben, Rueckert, Salli,
  Suetens, Wang, Wilms, Kirschke, Krämer, Münte, Schramm, Wiest, Handels, and
  Reyes]{Maier2017}
Oskar Maier, Bjoern~H. Menze, Janina von~der Gablentz, Levin Häni, Mattias~P.
  Heinrich, Matthias Liebrand, Stefan Winzeck, Abdul Basit, Paul Bentley, Liang
  Chen, Daan Christiaens, Francis Dutil, Karl Egger, Chaolu Feng, Ben Glocker,
  Michael Götz, Tom Haeck, Hanna-Leena Halme, Mohammad Havaei, Khan~M.
  Iftekharuddin, Pierre-Marc Jodoin, Konstantinos Kamnitsas, Elias Kellner,
  Antti Korvenoja, Hugo Larochelle, Christian Ledig, Jia-Hong Lee, Frederik
  Maes, Qaiser Mahmood, Klaus~H. Maier-Hein, Richard McKinley, John Muschelli,
  Chris Pal, Linmin Pei, Janaki~Raman Rangarajan, Syed M.~S. Reza, David
  Robben, Daniel Rueckert, Eero Salli, Paul Suetens, Ching-Wei Wang, Matthias
  Wilms, Jan~S. Kirschke, Ulrike~M. Krämer, Thomas~F. Münte, Peter Schramm,
  Roland Wiest, Heinz Handels, and Mauricio Reyes.
\newblock {ISLES} 2015 - {A} public evaluation benchmark for ischemic stroke
  lesion segmentation from multispectral {MRI}.
\newblock \emph{Medical Image Analysis}, 35:\penalty0 250--269, January 2017.
\newblock ISSN 1361-8415.
\newblock \doi{10.1016/j.media.2016.07.009}.

\bibitem[Menze et~al.(2015)Menze, Jakab, Bauer, Kalpathy-Cramer, Farahani,
  Kirby, Burren, Porz, Slotboom, Wiest, Lanczi, Gerstner, Weber, Arbel, Avants,
  Ayache, Buendia, Collins, Cordier, Corso, Criminisi, Das, Delingette,
  Demiralp, Durst, Dojat, Doyle, Festa, Forbes, Geremia, Glocker, Golland, Guo,
  Hamamci, Iftekharuddin, Jena, John, Konukoglu, Lashkari, Mariz, Meier,
  Pereira, Precup, Price, Raviv, Reza, Ryan, Sarikaya, Schwartz, Shin, Shotton,
  Silva, Sousa, Subbanna, Szekely, Taylor, Thomas, Tustison, Unal, Vasseur,
  Wintermark, Ye, Zhao, Zhao, Zikic, Prastawa, Reyes, and Leemput]{Menze2015}
B.~H. Menze, A.~Jakab, S.~Bauer, J.~Kalpathy-Cramer, K.~Farahani, J.~Kirby,
  Y.~Burren, N.~Porz, J.~Slotboom, R.~Wiest, L.~Lanczi, E.~Gerstner, M.~A.
  Weber, T.~Arbel, B.~B. Avants, N.~Ayache, P.~Buendia, D.~L. Collins,
  N.~Cordier, J.~J. Corso, A.~Criminisi, T.~Das, H.~Delingette, Ç. Demiralp,
  C.~R. Durst, M.~Dojat, S.~Doyle, J.~Festa, F.~Forbes, E.~Geremia, B.~Glocker,
  P.~Golland, X.~Guo, A.~Hamamci, K.~M. Iftekharuddin, R.~Jena, N.~M. John,
  E.~Konukoglu, D.~Lashkari, J.~A. Mariz, R.~Meier, S.~Pereira, D.~Precup,
  S.~J. Price, T.~R. Raviv, S.~M.~S. Reza, M.~Ryan, D.~Sarikaya, L.~Schwartz,
  H.~C. Shin, J.~Shotton, C.~A. Silva, N.~Sousa, N.~K. Subbanna, G.~Szekely,
  T.~J. Taylor, O.~M. Thomas, N.~J. Tustison, G.~Unal, F.~Vasseur,
  M.~Wintermark, D.~H. Ye, L.~Zhao, B.~Zhao, D.~Zikic, M.~Prastawa, M.~Reyes,
  and K.~Van Leemput.
\newblock The multimodal brain tumor image segmentation benchmark (brats).
\newblock \emph{IEEE Transactions on Medical Imaging}, 34\penalty0
  (10):\penalty0 1993--2024, Oct 2015.
\newblock ISSN 0278-0062.
\newblock \doi{10.1109/TMI.2014.2377694}.

\bibitem[Noh et~al.(2015)Noh, Hong, and Han]{Noh2015}
Hyeonwoo Noh, Seunghoon Hong, and Bohyung Han.
\newblock Learning deconvolution network for semantic segmentation.
\newblock In \emph{The IEEE International Conference on Computer Vision
  (ICCV)}, December 2015.

\bibitem[Peng et~al.(2017)Peng, Zhang, Yu, Luo, and Sun]{Peng2017}
Chao Peng, Xiangyu Zhang, Gang Yu, Guiming Luo, and Jian Sun.
\newblock Large kernel matters -- improve semantic segmentation by global
  convolutional network.
\newblock In \emph{The IEEE Conference on Computer Vision and Pattern
  Recognition (CVPR)}, July 2017.

\bibitem[Pereira et~al.(2016)Pereira, Pinto, Alves, and Silva]{Pereira2016}
S.~Pereira, A.~Pinto, V.~Alves, and C.~A. Silva.
\newblock Brain {Tumor} {Segmentation} {Using} {Convolutional} {Neural}
  {Networks} in {MRI} {Images}.
\newblock \emph{IEEE Transactions on Medical Imaging}, 35\penalty0
  (5):\penalty0 1240--1251, May 2016.
\newblock ISSN 0278-0062.
\newblock \doi{10.1109/TMI.2016.2538465}.

\bibitem[Randhawa et~al.(2016)Randhawa, Modi, Jain, and Warier]{Randhawa2016}
Ramandeep~S. Randhawa, Ankit Modi, Parag Jain, and Prashant Warier.
\newblock \emph{Improving {Boundary} {Classification} for {Brain} {Tumor}
  {Segmentation} and {Longitudinal} {Disease} {Progression}}.
\newblock Brainlesion: {Glioma}, {Multiple} {Sclerosis}, {Stroke} and
  {Traumatic} {Brain} {Injuries}. Springer International Publishing, Cham,
  2016.
\newblock ISBN 978-3-319-55524-9.
\newblock DOI: 10.1007/978-3-319-55524-9\_7.

\bibitem[Ronneberger et~al.(2015)Ronneberger, Fischer, and
  Brox]{Ronneberger2015}
Olaf Ronneberger, Philipp Fischer, and Thomas Brox.
\newblock U-net: Convolutional networks for biomedical image segmentation.
\newblock In Nassir Navab, Joachim Hornegger, William~M. Wells, and
  Alejandro~F. Frangi, editors, \emph{Medical Image Computing and
  Computer-Assisted Intervention -- MICCAI 2015}, pages 234--241, Cham, 2015.
  Springer International Publishing.
\newblock ISBN 978-3-319-24574-4.

\bibitem[Schmidt et~al.(2012)Schmidt, Gaser, Arsic, Buck, Förschler, Berthele,
  Hoshi, Ilg, Schmid, Zimmer, Hemmer, and Mühlau]{Schmidt2012}
Paul Schmidt, Christian Gaser, Milan Arsic, Dorothea Buck, Annette Förschler,
  Achim Berthele, Muna Hoshi, Rüdiger Ilg, Volker~J. Schmid, Claus Zimmer,
  Bernhard Hemmer, and Mark Mühlau.
\newblock An automated tool for detection of {FLAIR}-hyperintense white-matter
  lesions in {Multiple} {Sclerosis}.
\newblock \emph{NeuroImage}, 59\penalty0 (4):\penalty0 3774--3783, February
  2012.
\newblock ISSN 1053-8119.
\newblock \doi{10.1016/j.neuroimage.2011.11.032}.

\bibitem[Song et~al.(2016)Song, Chou, Chen, Huang, and Liu]{Song2016}
Bi~Song, Chen-Rui Chou, Xiaojing Chen, Albert Huang, and Ming-Chang Liu.
\newblock Anatomy-guided brain tumor segmentation and classification.
\newblock In Alessandro Crimi, Bjoern Menze, Oskar Maier, Mauricio Reyes,
  Stefan Winzeck, and Heinz Handels, editors, \emph{Brainlesion: Glioma,
  Multiple Sclerosis, Stroke and Traumatic Brain Injuries}, pages 162--170,
  Cham, 2016. Springer International Publishing.
\newblock ISBN 978-3-319-55524-9.

\bibitem[Yushkevich et~al.(2006)Yushkevich, Piven, Cody~Hazlett, Gimpel~Smith,
  Ho, Gee, and Gerig]{Yushkevich2006}
Paul~A. Yushkevich, Joseph Piven, Heather Cody~Hazlett, Rachel Gimpel~Smith,
  Sean Ho, James~C. Gee, and Guido Gerig.
\newblock User-guided {3D} active contour segmentation of anatomical
  structures: Significantly improved efficiency and reliability.
\newblock \emph{Neuroimage}, 31\penalty0 (3):\penalty0 1116--1128, 2006.

\end{thebibliography}

\end{document}